\documentclass{article}

\usepackage{outlines}
\usepackage[disable]{todonotes}
\usepackage{amsmath}
\usepackage{amssymb}
\usepackage{wrapfig}
\usepackage{multirow}
\usepackage{algorithm}
\usepackage{amsthm}
\newtheorem{theorem}{Theorem}[]

\usepackage{xcolor}
\usepackage{algpseudocode}
\usepackage[font=footnotesize]{caption}
\usepackage{subcaption}
\def\subparagraph{} 

\usepackage{pgfplotstable}
\usepackage{ifthen}
\pgfplotsset{
compat=newest,
width=0.3\textwidth,
}
\usepackage{arydshln}
\usepackage[final]{corl_2021} 

\title{Learning Model Preconditions for \\Planning with Multiple Models }
\newcommand{\tf}[2]{^{\mathrm{#2}}T_{#1}}
\newcommand{\aincomplete}{A_{inc}}
\newcommand{\subfigandcaption}[3]{
    \begin{subfigure}[b]{#3\textwidth}
    \includegraphics[width=\textwidth]{#2}
    \vspace{-14pt}
    \caption{\scriptsize{#1}}
    \end{subfigure}%
    \hspace{-2pt}
}

\newcommand{\dmax}{d_{\mathrm{max}}}

\newcommand{\ed}[1]{{#1}}
\DeclareMathOperator*{\argminC}{\arg\min}
\newcommand{\open}{\text{OPEN}}
\newcommand{\closed}{\text{CLOSED}}
\newcommand{\reducesubtitlespace}{\vspace{-0.3cm}}
\newcommand{\reducesectionspace}{\vspace{-0.3cm}}
\author{
  Alex LaGrassa\\
  Robotics Institute\\
  Carnegie Mellon University\\
  \texttt{alagrass@andrew.cmu.edu} 
  \And
  Oliver Kroemer\\
  Robotics Institute\\
  Carnegie Mellon University\\
  \texttt{okroemer@andrew.cmu.edu} \\
}

\begin{document}
\maketitle


\begin{abstract}
Different models can provide differing levels of fidelity when a robot is planning. Analytical models are often fast to evaluate but only work in limited ranges of conditions. Meanwhile, physics simulators are effective at modeling complex interactions between objects but are typically more computationally expensive. Learning when to switch between the various models can greatly improve the speed of planning and task success reliability. In this work, we learn model deviation estimators (MDEs) to predict the error between real-world states and the states outputted by transition models. MDEs can be used to define a model precondition that describes which transitions are accurately modeled. We then propose a planner that uses the learned model preconditions to switch between various models in order to use models in conditions where they are accurate, prioritizing faster models when possible. We evaluate our method on two real-world tasks: placing a rod into a box and placing a rod into a closed drawer.
\end{abstract}
\vspace{-0.3cm}
\section{Introduction}
\vspace{-0.3cm}

Predictive models that are helpful for intelligent behaviors can take a variety of different forms, e.g. analytical models~\citep{goyal1991planar, pan2020decision}, physics simulations~\citep{liang2018gpu, coumans2021}, and learned models~\citep{hafner2019learning, sundaralingam2021hand, nagabandi2020deep}, and the best choice of model is context-dependent. For example, a high fidelity simulator can model complex interactions between a large number of objects when dumping a pile of non-convex objects on a table~\citep{macklin2014unified}. However, for a simple tabletop pick and place task, a simulator model is often unnecessary when a simple kinematic model is available.
\begin{figure}[H]
    \centering
    \vspace{3pt}
    \includegraphics[width=0.91\textwidth]{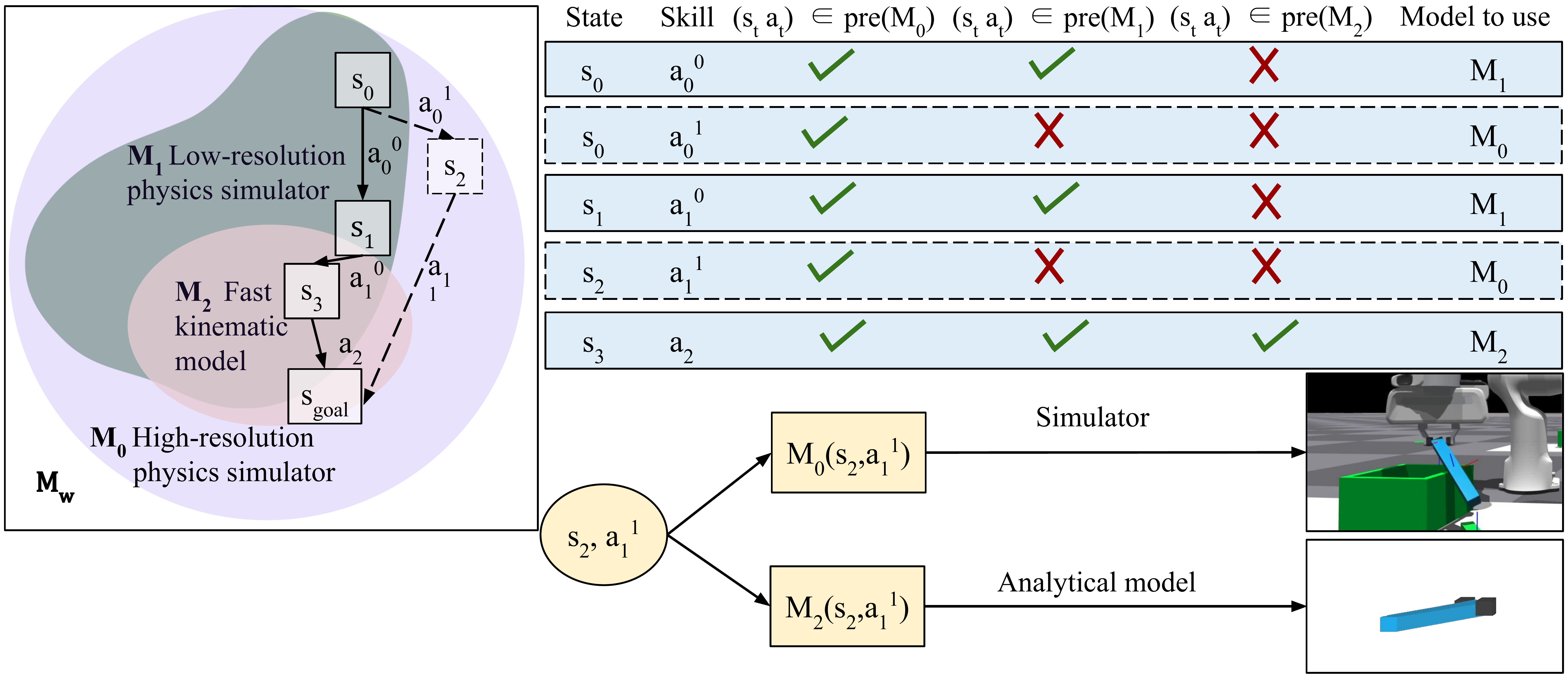}
    \caption{ \footnotesize Each colored region on the left represents a model precondition, which we learn in this work for analytical and simulator models for manipulation skills. In this example, a kinematic model is faster than a low-fidelity simulator, and a low-fidelity simulator is faster than a high-fidelity simulator. On the right we show for each step which state-action pair $(s_t, a_t^i)$ was in the precondition for each model, and the best model to use to evaluate that edge. The states and actions on the path chosen for the plan have a bold border. States not expanded because of high model cost have a dashed border. Multiple actions at a state $s_t$ are denoted by superscripts: For example, two different actions from $s_0$ would be $[a^{0}_0, a^1_0]$.}
    \label{fig:modelpreconditions}
\end{figure}
A planner may need to make many predictions to find a plan, so it is important to only use  computationally expensive models when they are most needed. \ed{We define \textit{model preconditions}, shown abstractly in Figure~\ref{fig:modelpreconditions}, that demarcate regions of states and actions where the model accurately represents real-world dynamics. We use those model preconditions during planning to leverage the complementary advantages and disadvantages of different models.}
\todo[inline]{It would be helpful to add something that describes the benefits of reasoning about model error at a global area to avoid reaching states where model error is expected, which would also make it more clear what benefits our method would have that MPC does not account for. }

\ed{To capture the model preconditions, the robot learns to predict the total state deviation as a continuous value for each model. These deviation predictions are used to inform the planner to avoid transitions that are inaccurately modeled. The amount of acceptable model error is task-dependent, but by predicting state deviation as a continuous value rather than a binary one, the thresholds can be set based on the task accuracy requirements. }


This work's contributions include a method for defining model preconditions using a learned scalar model deviation term and then using those preconditions for planning with multiple models to minimize model error at execution time. 
We first show how to fit Model Deviation Estimators (MDEs) from planning-relevant data to predict the error between a transition model and real-world dynamics. We then use MDEs to define the model preconditions for multiple skills for specific tasks.
\ed{We evaluate our method for learning and planning with MDEs on two real-world tasks: placing a rod that can rotate in-hand into a box and placing a rod into a closed drawer. The robot has access to incomplete analytical models and a simulator model for each action class. Like in many real-world planning applications, none of the models match the real-world dynamics for all possible interactions. We compare two planning approaches with MDEs against other model selection baselines. } 
\reducesectionspace
\section{Related work}
\reducesectionspace
We review related work relevant to our two main contributions: predicting model error from data and planning using multiple imperfect models by reasoning about planning speed and reliability tradeoffs. 

\textbf{Improving planning reliability with past experience:} Learning model preconditions for planning draws from the high-level idea of identifying similarities to states encountered in previous plans to apply behaviors that previously led to successful plans, which can significantly improve planning speed~\citep{kim2019learning,phillips2012graphs}. Plan reliability can also be improved by learning skill preconditions, which describe the states from which skills achieve desired effects with high probability~\cite{konidaris2018from, Sharma-2020-126757, wang2021learning}. Instead of using skill preconditions, we use transition model preconditions.

\textbf{Avoiding regions with model error:} For predicting where the model is accurate using data,~\citet{Chou2021ModelEP} and~\citet{knuth2021planning} use hyperspheres to represent proximity to training data for a learned model as a proxy for the model precondition. \textsc{Cmax} and \textsc{Cmax}++ also use hyperspheres, but use points where the action effects differed significantly from those predicted by the internal model~\citep{Vemula-2020-122905, vemula2020cmax++}. \ed{However, proximity to training data and predicted collisions do not always capture model accuracy when executed on a real robot. Our work proposes a method to fit an estimator from plan execution data to predict whether model error will occur. Other works bias search away from states where the simple planning model is known to be inaccurate, but also do not use other models to compensate for transitions the simple model cannot model accurately, while our planner uses a slower model where appropriate ~\citep{knuth2021planning, Chou2021ModelEP, Vemula-2020-122905, mcconachie2020learning}. This distinction of not using alternative models when model error is predicted is especially significant when no one model can be used to compute a plan accurately, such as when each model only reasons about a subset of interactions necessary to achieve a task.}

\ed{\textbf{Predicting model error:} 
The works most similar to ours use visual similarity to model the uncertainty of their simple model in~\citet{power2021keep} and learning a classifier in the work of~\citet{Mitranoeabd8170} and~\citet{mcconachie2020learning}. The primary difference is that they do not use their model error estimates as criteria for switching models, whereas we use the planning data to learn model error estimates to decide between multiple models. \citet{McConachie2020bandit} does use multiple models, but uses task progress rather than error as selection criteria and does not do multi-step planning.
Additionally, these works assume access to a simulator accurate enough to compute the ground-truth labels for accurate model predictions. We relax that assumption by treating the simulator as an additional model, which can also exhibit model error. Our labels then come from the real world. 
The methods above are also focused on a rope manipulation domain, where errors can usually be corrected with a recovery policy, as shown in~\citep{Mitranoeabd8170}. Many actions in the manipulation domain in which we evaluate our model error prediction lead to unrecoverable states, such as an object lying in an un-graspable pose in a constrained space. }

\textbf{Planning with multiple models: } The work of \citet{saleem2020planning} uses a rule where a simulator is only used when collision is detected using an internal model to speed up planning. \ed{ However, there are many situations where there are transitions in which objects are in contact but an analytical model can still be used. Our planning algorithm builds on Multi-Graph Multi-Heuristic A*~\citep{youakim2018motion}, which presents an algorithm that selects between models which operate on different parts of the state. The models we consider in this work operate on the same objects in the state, which makes choosing a model not straightforward. In Multi-Resolution A*~\cite{du2020multi}, lower resolution actions are used to make planning faster, but are assumed to all be modeled correct. Furthermore, for the models we use, the best model to use depends on both the state and action rather than just state. }

\reducesectionspace
\section{Problem Formulation}
\reducesectionspace
The overall problem is to first learn the MDEs to predict the total model deviation given a state, action, and transition model, then use MDEs for planning using multiple models, prioritizing planning using the faster models where possible while still minimizing cost. 

The state space is denoted by $\mathcal{S}$ and is assumed to be fully observable at the end of an action. 
The action space $\mathcal{A}$ is a set of closed-loop skills $a \in \mathcal{A}$, parameterized by $\theta$ which are available to the robot. A user-defined skill-specific parameter generation function generates potentially useful parameters given a state. Skills follow the options formulation \cite{sutton1999between}.
Skills are executed using a low-level controller until a termination condition is reached. The actions are intended to be run for a significant period of time such that feedback control at a lower level of abstraction can correct for low-level errors, leaving the MDEs to capture model error that is harder to correct. 
Skills have an associated precondition set, $\mathrm{pre}(a) \subseteq \mathcal{S}$ describing the set of states where the skill can be executed.
The skill preconditions narrow the space where the skill has defined behavior, but do not comprehensively describe the set of states where the dynamics models are accurate. The skill precondition simply relates to the skill's applicability and are separate from MDEs and models of the skills' effects, which we describe next. 

The robot is given an ordered list of models $[M_0, M_1, \ldots ,M_{K}]$, used to compute the forward dynamics: $\hat{s}' \gets M_i(s, a)$. \ed{The models are ordered by increasing computation speed, where $M_i$ is slower to evaluate than $M_j$ if $i < j$}. For example, $M_0$ may be a high-fidelity simulator that is slow but very accurate, $M_1$ a simulator with a coarse timestep discretization, $M_2$ an analytical model requiring more computation, and $M_3$ a simple linear model.
It is preferable to use the model with the highest $i$, corresponding to the fastest model, that is accurate for that $(s, a)$. 
To measure model accuracy, the robot is given a distance function between states $d(s_i, s_j)$, such as Euclidean distance. The MDE predicts $\hat{d}$ from $\phi(s)$ and $a$, where $\phi(s)$ is a function that extracts relevant features \ed{such as object poses or distances between objects}. 

Lastly, the robot is given a task that includes a goal set, a cost function, and optionally additional task-specific action parameter generators. The state satisfies the goal if it is in the goal set $\mathcal{S}^g$. The cost function describes the cost of high-level state transitions and is denoted by $c(s, a, s')$. 

To train the MDEs, the robot needs to first collect an offline dataset $\mathcal{D}$ of real-world transitions $(s, a, s')$. We assume the robot knows the initial state distribution but not all states visited during planning. Initial states are sampled from the initial state distribution, but the rest of the states are determined from planning. Then the robot trains the MDE $\hat{d}(s, a)$ which estimates $d({s'}, \hat{s}')$ using $\mathcal{D}$. The eventual goal is for the robot to search for a high-level plan of actions $[a_0, a_1, \ldots, a_{T-1}]$ to create a sequence of states $[s_0, s_1, \ldots, s_{T}]$ such that the final state is a goal state: $s_{T} \in \mathcal{S}^g$.

\reducesectionspace
\section{Approach}
\reducesectionspace
First we explain how to define and learn model preconditions using MDEs.
We then show how MDEs can be used in planning to use multiple models while prioritizing states that can be reliably evaluated using simpler models. 
\reducesubtitlespace
\subsection{Learning Model Preconditions Using Model Deviation Estimators}
\reducesubtitlespace
\label{sec:learningmodelpreconds}

If a transition is in a model's precondition, then that model can be used to accurately model that transition. A model precondition is defined using MDEs as:
\setlength{\belowdisplayskip}{2pt}
\setlength{\abovedisplayskip}{2pt}
\begin{equation}
\mathrm{pre}(M_i) = \{(s, a) \, | \, \hat{d}(M_i( s, a), s') < \dmax\}
\label{eq:modelpre}
\end{equation}
Training data for MDEs is collected by computing and executing plans using all models. If a path to the goal is found, it is executed. Each transition is saved as $(s, a, s')$ tuples for every transition observed in the path. The list of transitions is expensive to collect as it requires real-world data, but several datasets for MDEs can be derived from it. 

To train the MDE for a particular model $M_i$, $\hat{s}'$ is computed using $M_i$ for each $(s, a)$ in $\mathcal{D}$. The labels are $d(s', \hat{s}')$. Training data for transitions is used across all models even though that transition was computed using just one model during planning. 

The inputs are state features, $\phi(s)$ concatenated with $\theta$. $\phi(s)$ can be the identity function, but local features such as the distances between objects can enable easier generalization.

We train an MDE for every combination of skill and model, though MDEs can be shared if skill parameters have similar meaning. The reason why MDEs are regression models instead of classification models is to enable sharing of MDEs across tasks with different accuracy requirements. \ed{To model the MDEs we use a 3-layer MLP with 32 hidden units in each layer. }

Next we describe the loss function for MDEs. An \textit{underestimated} predicted deviation is worse for plan reliability than an \textit{overestimated} predicted deviation. Overestimates cause the planner to be overly conservative, but an underestimate can cause plans to be executed unreliably on the real robot. As a result, we propose a loss function that penalizes underestimates more than overestimates: $L_g(d, \hat{d}) =  c_1\mathrm{max}(0, d-\hat{d})^2 + c_2 \mathrm{max} (0, \hat{d} - d)^2$. To get the desired behavior, we set $c_1 > c_2$.  
\reducesubtitlespace
\subsection{Planning With Multiple Models Using MDEs}
\reducesubtitlespace
Model preconditions can be used in the same manner as skill preconditions by using the model preconditions as a constraint during planning, such as by using Equation~\ref{eq:modelpre} as an inequality constraint during optimization, where the $M$ used at each $t$ are additional variables. Model preconditions as predicate constraints can be applied to STRIPS-style planners using actions with a particular $M$ to compute skill effects by including $(s, a) \in \mathrm{pre}(M)$ to the existing action preconditions as an additional predicate. The simplest possible way to speed up planning using model preconditions is to apply the fastest model that satisfies the precondition, which we call \textit{Prioritized Selection} (PS). 

Although MDEs can be used to select between models in a more general setting, we propose a planner that can more effectively use MDEs by using a user-defined model priority weighting to prioritize expanding subgraphs that are less expensive to evaluate. We call this technique Prioritized Expansion (PE). Using PE, our planner prioritizes simpler models by using a weighting factor inspired by the one used in Multi-Heuristic A*(MHA*)~\cite{aine2016multi}. Because our planner uses successors from different models, we also build on Multi-Heuristic Multi-Representation A* ~\cite{youakim2018motion} and Multi-Resolution A* (MRA*)~\cite{du2020multi}.

\ed{Our proposed planning algorithm using both PS and PE uses up to $K$ models for each skill. Models can be shared across skills. The planner expands $K$ implicit bidirectional graphs, which maintain their own open queues for nodes to expand and closed sets for nodes that have already been expanded, but all graphs share successors. All open queues are initialized with the start state and all closed sets are initialized to $\emptyset$. 

An anchor search expands its graph by selecting between all models, and the $K-1$ additional graphs expand using the $K-1$ faster models. Priorities between models are adjustable using a list of model preference weights, $w = [w_0, w_1, \ldots, w_{K-1}]$ decreasing in value representing how much to penalize slower models.} $\frac{w_0}{w_{K-1}} \epsilon$ also upper bounds the sub-optimality incurred by using faster models, which is shown in Appendix~\ref{sec:proof}.

We now describe how the separate queues and graphs are used to switch between models. 

Similar to standard Weighted A* (WA*), each planning node has a cost $g$, heuristic value $h$, and priority $f = g + \epsilon h$. The function $\textsc{MinKey}()$ returns the node in a queue with the lowest $f$ value. \ed{To handle partial expansions, a node can optionally have a set of un-evaluated actions associated with it, $\aincomplete(s)$.} 
\ed{The anchor search uses open queue $\open_0$ and closed set $\closed_0$. It uses a \textit{full expansion}, which computes the successor with the fastest model that satisfies the precondition for a particular state and action.} If no model satisfies the precondition, there is no successor for that transition. 
\ed{Nodes that are fully expanded in the anchor search are added to all closed sets.} \ed{Each additional search maintains its own open queue $\open_i$ and closed set $\closed_i$. It computes successors using a \textit{partial expansion} only using $M_i$ as long as $(s, a) \in \mathrm{pre}(M_i)$, saving other actions for expansion by other graphs. }

\textbf{Graph and node selection:} We expand from queue $i$ using a rule inspired from MHA*:
\setlength{\belowdisplayskip}{2pt}
\setlength{\abovedisplayskip}{2pt}
\begin{equation}
\ed{i = \argminC_{i < K} \; w_i \, \open_i.\textsc{MinKey}()}
\label{eq:expansionrule}
\end{equation} 
If $i = 0$, then a node is expanded in the anchor search using the full expansion. If $i > 0$ then the node is expanded using a partial expansion for graph $i$. The node to be expanded is $\open_i.\textsc{MinKey}()$, as with standard WA* search. 

\begin{figure}
    \centering
    \includegraphics[width=\textwidth]{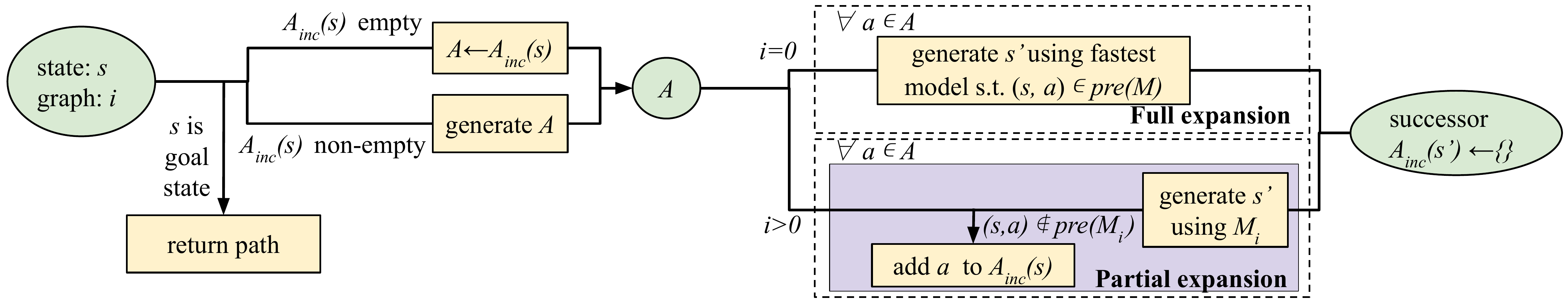}
    \caption{\footnotesize Graphical representation of the two types of expansions using model preconditions. Green ovals are data and yellow rectangles represent operations on data. The left side shows how actions are selected for both expansion types. The right side shows full expansion (top) and partial expansion (bottom). }
    \label{fig:plannergraphic}
    \vspace{-0.5cm}
\end{figure}

We begin by describing steps used for both types of expansions, and then describe what is different. All expansions begin by first checking the goal condition. As shown in Figure~\ref{fig:plannergraphic}, the search is terminated when a goal state is expanded. Otherwise, expansion begins by selecting the set of actions to evaluate, $A$. $A$ is $\aincomplete$ if $\aincomplete$ is non-empty. If not, a parameter generation function generates a candidate list of possible parameterized actions. The graph to expand is chosen using the rule in Equation~\ref{eq:expansionrule}. For each successor $s'$, if the path found to $s'$ is lower cost than any current path found to $s'$ in any graph, the path is updated using $s'$. Then $s'$ is added to to all $\open_i$ if it is not in $\closed_i$. 

\textbf{Full expansion:}
Full expansion is used as a fallback when the other searches with faster models run out of nodes or only contain highly suboptimal nodes.
\ed{As shown in the top right box in Figure~\ref{fig:plannergraphic}, full expansion uses PS by evaluating all successors using $M_i(s, a)$ for the highest value of $i$ that satisfies $(s, a) \in \mathrm{pre}(M_i)$. } Then, $s$ is added to all closed sets. 

\textbf{Partial expansion:}
\ed{The intuition behind using partial expansion in PE is to evaluate nodes using faster models sooner while delaying slower evaluations. For all $(s, a) \in \mathrm{pre}(M_i)$,
the successors $s'$ are computed using $M_i$, and they are added to all open queues. For all states and actions not in the precondition, the successors are not evaluated. Instead, those actions are added to $\aincomplete(s)$ for later evaluation in other graphs, then the node is added to $\closed_i$.
} This step helps bias the search to nodes that are cheap to evaluate, while still allowing the planner to reason about when to use more expensive models for completeness or optimality.

\vspace{-0.3cm}
\section{Experiments}
\vspace{-0.3cm}
\newcommand{\inserttask}{\texttt{RodInDrawer}}
\newcommand{\pickandplacetask}{\texttt{RodInBox}}

\newcommand{\planFoundAnalyticalOnly}{
(\inserttask, 0.0)
(\pickandplacetask, 1.0)}

\newcommand{\planFoundOurs}{
(\inserttask, 1.0)
(\pickandplacetask, 1.0)}

\newcommand{\planFoundHighFidSim}{
(\inserttask, 1.0)
(\pickandplacetask, 1.0)}

\newcommand{\planSuccessAnalyticalOnly}{
(\inserttask, 0)
(\pickandplacetask, 0.4)}

\newcommand{\planSuccessHighFidSim}{
(\inserttask, 0.4) 
(\pickandplacetask, 0.9)} 

\newcommand{\planSuccessOurs}{
(\inserttask, 0.5) 
(\pickandplacetask, 0.7)} 
\newcommand{\robotandrods}{\textit{Analytical (Pick \& Place)\,}}
\newcommand{\robotanddrawer}{\textit{Analytical (Drawer)\,}}
\newcommand{\simulator}{\textit{Simulator}}
\newcommand{\multqs}{(PS+PE)}
\newcommand{\oneq}{(PS only)}
First, we describe the metrics we use to evaluate our method, then describe the experimental setup we use for two real-world task domains shown in Figure~\ref{fig:experimentalsetup}: placing one of two steel rods in a box (\texttt{RodInBox}), and in a drawer (\texttt{RodInDrawer}). \ed{Then, we show the accuracy of MDEs for each skill. Finally, we evaluate two planners that use MDEs on two real-world tasks against baselines: one using a single queue with model preconditions as constraints and another using our multiple-queue planner described in Section 4.2. 

We compare timing metrics, demonstrate the model selection for each skill in the \texttt{RodInDrawer} task, and discuss how the model selection distribution in each algorithm impacts plan time and reliability. } For plan performance, we test: \emph{1)} the time to compute a plan (seconds), \emph{2)} the model evaluation rate (model evaluations per second) to measure the planner's ability to explore efficiently, \emph{3)} the success rate in computing a plan, and \emph{4)} the success rate achieving the goal in the real-world tasks if a plan was found (to measure reliability).


\textbf{Tasks:} 
\ed{For both \texttt{RodInBox} and \texttt{RodInDrawer} tasks, a Franka Emika Panda manipulates one of two steel rods into a desired container. The rod to place in the container is chosen randomly. Due to its weight, the rod can drop or rotate in the gripper (Figure~\ref{fig:failuremodes}). Both rods are placed in arbitrary reachable poses not occluded by other objects on the table.}

The first task is \texttt{RodInBox}. A state is in $\mathcal{S}^g$ if the target rod is in the green box shown in Figure~\ref{fig:experimentalsetup}. The purpose of this task is to evaluate what happens when our method is applied on a task where high precision is not necessary for task completion, but the effects of some actions are easier to model than others. The second task we evaluate on is \texttt{RodInDrawer}, shown in Figure~\ref{fig:experimentalsetup} where the goal is to place a rod in the partially closed drawer. Goal states are those where the rod is within the bounds of the drawer. Heuristics for both tasks are in Appendix~\ref{alg:planningAlg}.

Actions are movements in Cartesian end-effector space, implemented in the real world using the library from~\citet{zhang2020modular} and with end-effector attractors in simulation. In simulation, the world is fully observable. In the real world, the rod poses are estimated using an the overhead camera. When a grasped object is occluded by the robot, an in-hand camera is used to confirm that the object is in the gripper, and the previously recorded pose \ed{is} used. This perception is sufficient for our use case because the skills with the most variability do not end with an object in the gripper.

\newcommand{\setupWidth}{0.24}
\begin{figure}
    \centering
    \subfigandcaption{\texttt{RodInBox} (Sim)}{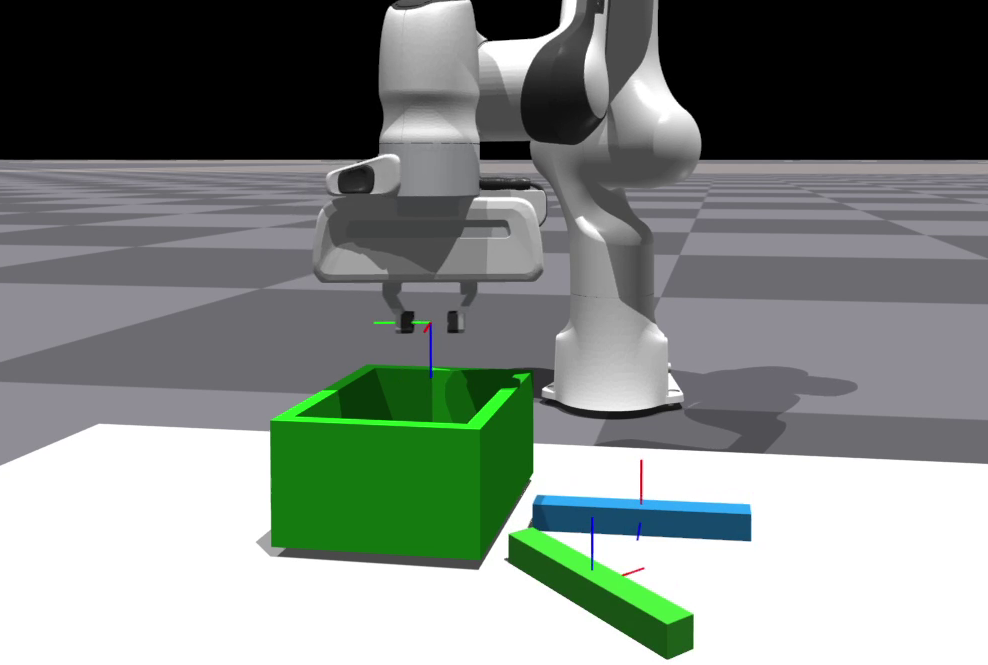}{\setupWidth}%
    \subfigandcaption{ \texttt{RodInBox} (Real)}{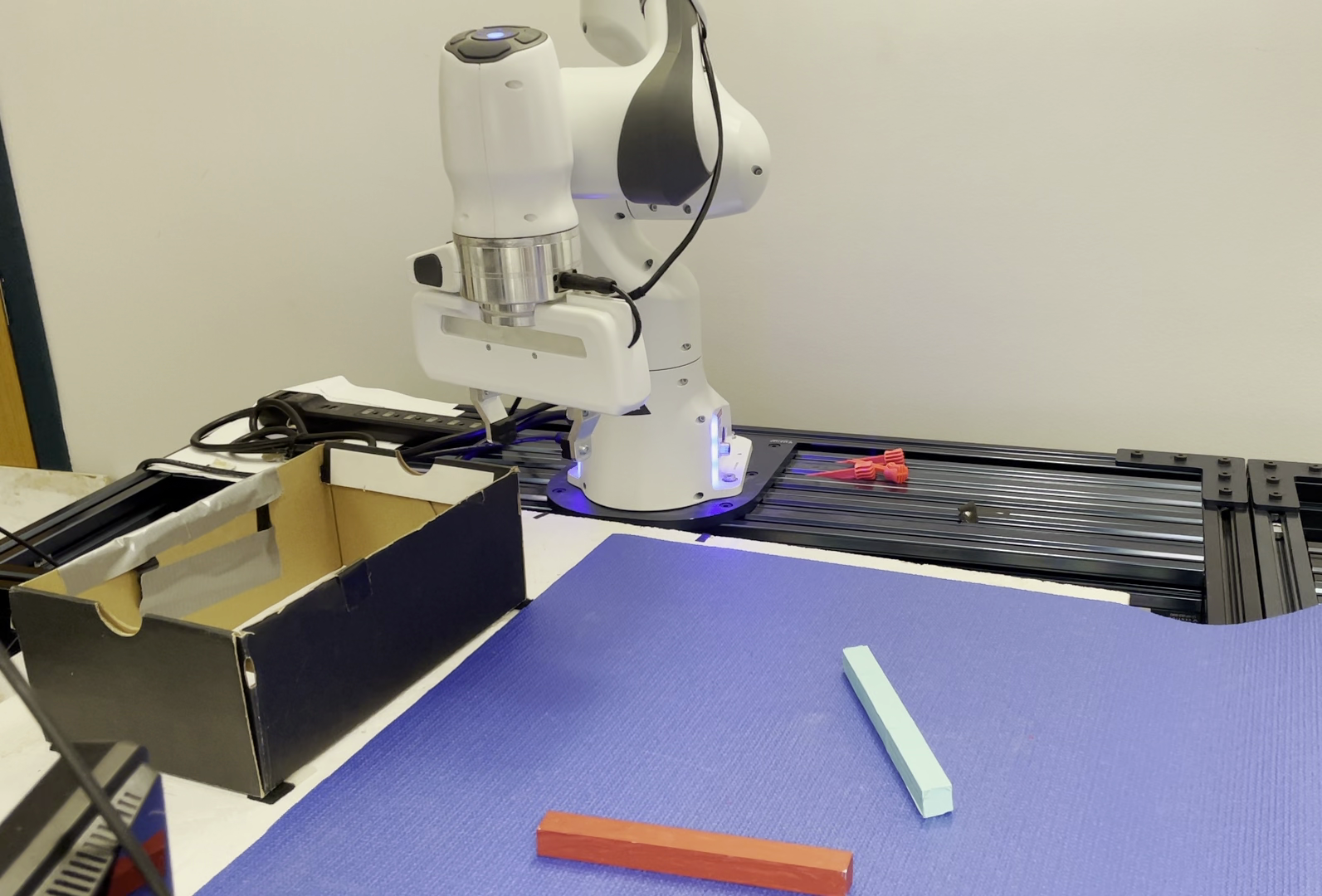}{\setupWidth} %
    \subfigandcaption{ \texttt{RodInDrawer} (Sim)}{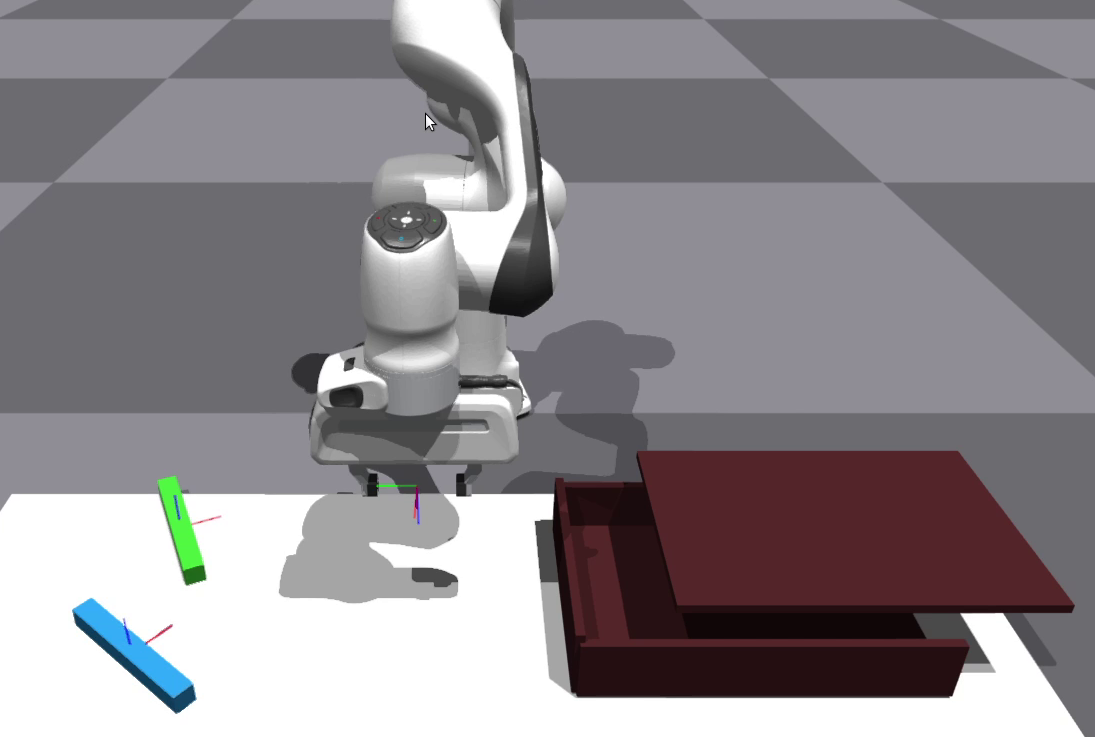}{\setupWidth}%
    \subfigandcaption{ \texttt{RodInDrawer} (Real)}{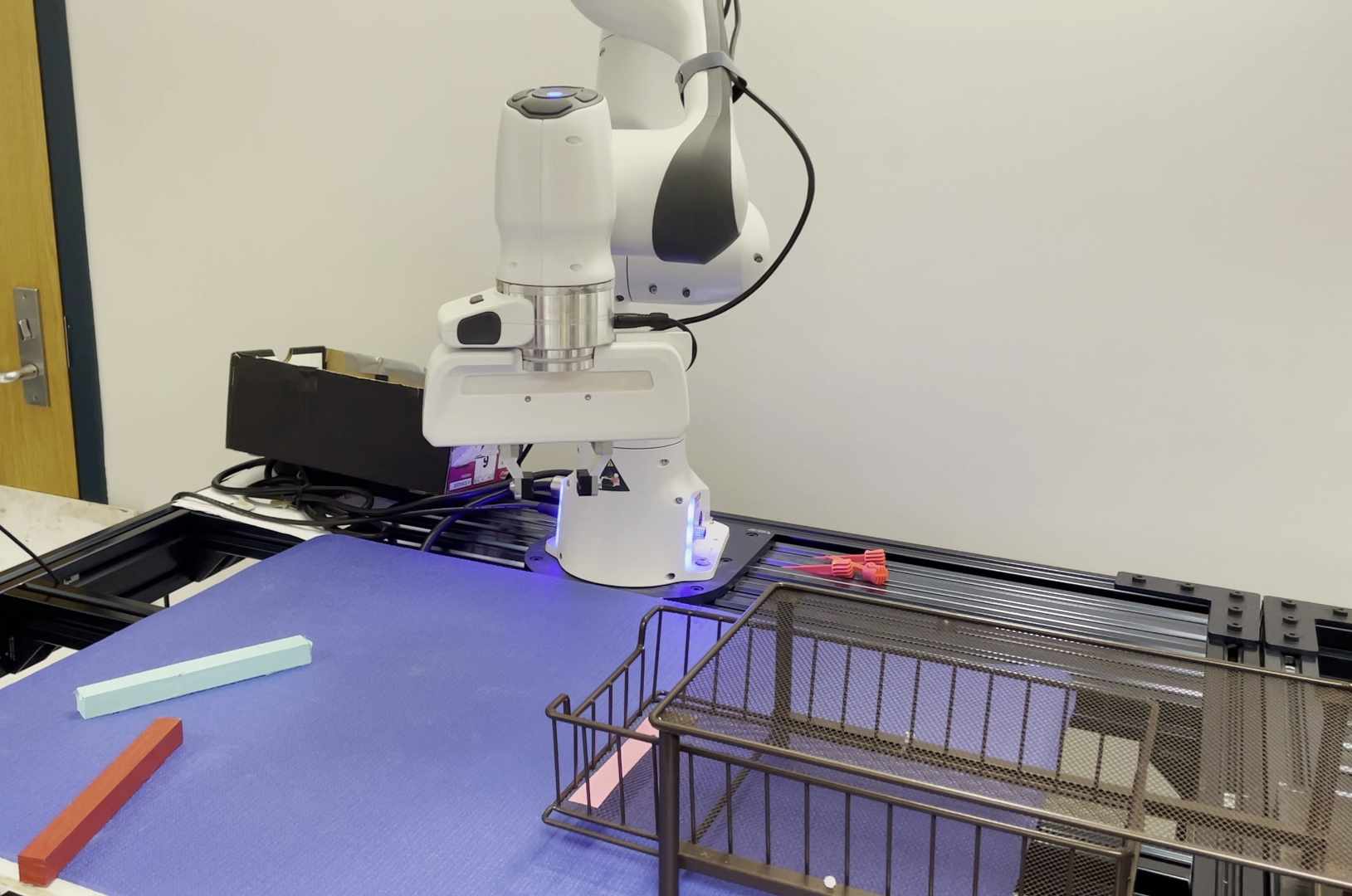}{\setupWidth}
    \caption{Robot setup for test tasks in simulation and on the real robot. 
    }
    \label{fig:experimentalsetup}
    \vspace{-0.5cm}
\end{figure}
\textbf{Skills: } The skills the robot uses are \textit{Pick}, \textit{LiftAndDrop}, and \textit{OpenDrawer}. Because the effect models used are high-level and the end-effector moves horizontally at a height above all obstacles, collisions are only checked at the beginning and end of each motion. The state includes the pose of all objects in the scene, including the robot end-effector pose.
\textit{Pick} is parameterized with the goal pose, which is sampled around the target rod. 
\textit{LiftAndDrop} brings the end-effector and rod to a target location, then opens the grippers. The precondition of \textit{LiftAndDrop} is that an object is between the grippers and that the goal pose will not cause collisions. 
\textit{OpenDrawer} moves the gripper to a fixed location relative to the front of the drawer, then opens the drawer moving to a final pose. Detailed descriptions of all the skills, including hyperparameters, are in Appendix~\ref{sec:skill_descs}.

\textbf{Models: } The robot has access to a simulator and two analytical models. One analytical model, \robotandrods, only computes the transition model for pick and place actions, assuming a rod close to the gripper is rigidly attached. When the rod is placed, it falls to a height determined by the object (if any) directly below it. Another model, \robotanddrawer, only reasons about a simplified articulation mechanism of the drawer, assuming the drawer edge is rigidly attached to the gripper if both surfaces are close, but does not model interactions between the rods and drawer. Details about each analytical model are in Appendix~\ref{sec:analyticalmodels}

The simulator we use for both tasks is IsaacGym~\cite{liang2018gpu}. For \texttt{RodInBox}, we use a setup with a box shown in Figure~\ref{fig:experimentalsetup}(a). The simulation parameters used for both experiments are shown in Appendix ~\ref{sec:simparams}. For \texttt{RodInDrawer}, the drawer is modeled as a single drawer chest using cuboids, as shown in Figure~\ref{fig:experimentalsetup}(c). To evaluate the result of a skill from a state $s$, the simulator sets the state of the world to $s$, executes skill $a$ and returns the resulting state as $\hat{s}'$.

\newcommand{\scinot}[2]{#1 \cdot 10^{#2}}
\begin{wraptable}{r}{0.6\textwidth}
    \centering
    \footnotesize\addtolength{\tabcolsep}{-5pt}
    \begin{tabular}{c | c | c | c}
    Skill & Model &  MAE & d \\
    \hline
         \textit{Pick} & \robotandrods & \ed{${0.2}$} & 0.4 (0.4) \\
         \textit{Pick} & \robotanddrawer  & ${0.2}$ & 0.4 (0.4) \\
         \textit{Pick} & \simulator  & ${0.7}$ & 4.7 (9.5)  \\
         \hdashline
         \textit{LiftAndDrop} & \robotandrods  & ${1.9}$ & 8.3 (7.41)\\
         \textit{LiftAndDrop} & \robotanddrawer&  ${7.6}$ & 11.6 (8.0)  \\
         \textit{LiftAndDrop} & \simulator &  ${2.5}$ & 11.0 (10.7)\\
         \hdashline
         \textit{OpenDrawer} & \robotandrods   & $0.4$ & 16.0 (0.4) \\
         \textit{OpenDrawer} & \robotanddrawer  & ${0.6}$ & 2.1 (0.3) \\
         \textit{OpenDrawer} & \simulator & ${1.2}$ & 7.0 (0.5) \\
    \end{tabular}
    \caption{\footnotesize Mean absolute error (MAE) and $d$ (mean and standard deviation) in centimeters for each model/skill combination. }
    \label{tab:lgEachSkill}
\end{wraptable}

\reducesubtitlespace
\subsection{MDE Accuracy}
\reducesubtitlespace
In this section, we evaluate MDE accuracy on each skill used, which correlates with the variance of $d$. As shown in Table~\ref{tab:lgEachSkill}, deviation prediction error is highest for \textit{LiftAndDrop} because there is a wider range of possible outcomes. The predicted deviation error for the simulator model is low but not zero because predicting the precise deviation requires an accurate real-world dynamics model. For example, a poorly-grasped rod may drop in simulation earlier than it would on the real-robot system by dragging on the ground differently. Since \robotanddrawer does not model movement of the rods, the exact deviation depends on the state and parameters, but the error is always predicted to be high for \textit{LiftAndDrop}, which is sufficient information to guide planning. The deviation for \textit{Pick} using \robotanddrawer is low because modeling \textit{Pick} does not require any physical interactions between robots and rods, and the robot is not in contact with the drawer. Simulator deviation prediction error for \textit{Pick} is high because when a drawer is in the environment, the joints sometimes interact with the drawer, causing interactions that do not occur in the real world. 

\reducesubtitlespace
\subsection{Planning Speed}
\reducesubtitlespace
\label{sec:eval}
For planning, we evaluate the performance from ten different initial start states per task. \texttt{RodInBox} uses \textit{Pick} and \textit{LiftAndDrop}, sampling five parameter vectors per skill. \texttt{RodInDrawer} uses \textit{Pick}, \textit{LiftAndDrop}, and \textit{OpenDrawer}, with three parameters generated per skill. $\epsilon=5$ for \texttt{RodInBox} and $\epsilon=10$ for \texttt{RodInDrawer}. The model preference weights are [10,1] corresponding to [\simulator, \robotandrods] for \texttt{RodInDrawer} and [10,1.1,1] corresponding to [\simulator, \robotanddrawer, \robotandrods] for \texttt{RodInBox}. The cost function is given by the total end effector distance covered in the trajectory. For timing experiments shown in Table~\ref{tab:timingLiftAndDrop}, we allow the planner to run for at most 300 seconds (5 minutes). 

\ed{\textbf{Baselines:} We compare planning using model preconditions to planning with each individual model, and randomly selecting which model to use when computing each successor. Additionally, we perform an ablation test on our full method \multqs \, by removing the additional queues used for prioritized expansion and only performing full expansions using the anchor search, though still using prioritized selection \oneq.}

\label{sec:timing}
\newcommand{\dashhskip}{\;-\;}
    \begin{table}
        \centering
    \begin{subfigure}[b]{0.47\textwidth}
        \begin{center}
        \footnotesize\addtolength{\tabcolsep}{-5pt}
        \begin{tabular}{ p{4cm}| p{1.1cm}| p{1.5cm}}
         \large
             \footnotesize{Method}  & \footnotesize{Plan time} & \footnotesize{Model eval. per second} \\
            \hline
            Ours\dashhskip \multqs &  0.40 & 25.2 \\
            Ours\dashhskip \oneq &  8.98 & 1.13 \\
            Random &  40.00 & 0.25 \\
            \ed{\robotandrods only} & \ed{0.19} & \ed{113.58}  \\
            \simulator\,only & 31.23 & 0.5  \\
        \end{tabular}
        \end{center}
        \caption{\footnotesize \texttt{RodInBox} }
        \label{tab:timingLiftAndDrop}
    \end{subfigure}%
        \hspace{0.025\textwidth}
        \begin{subfigure}[b]{0.47\textwidth}
        \ed{
        \footnotesize \addtolength{\tabcolsep}{-5pt}
        \begin{tabular}{ p{4cm}| p{1.1cm}| p{1.5cm}}
             \footnotesize{Method} & \footnotesize{Plan time} & \footnotesize{Model eval. per second} \\
            \hline
            Ours\dashhskip \multqs &  3.69 & 19.0\\
            Ours\dashhskip \oneq &  5.27 & 6.0\\
            Random & 44.36 & 0.6 \\
            \simulator\,only & 103.34 & 0.5  \\ 
             &  &  \\
        \end{tabular}
        }
        \vspace{.11cm}
        \caption{\footnotesize \texttt{RodInDrawer} }
        \label{tab:timingOpenDrawer}
        \end{subfigure}
        \caption{\footnotesize Planning times for methods that can find plans. We show the average time to compute a plan (not including timeouts).}
    \end{table}

For \texttt{RodInBox}, (Table~\ref{tab:timingLiftAndDrop}) the planning time and model evaluation rate are fastest using only the analytical model. Random model selection and simulator-only planning are the slowest. Because our method uses the fastest reliable model wherever possible, and for this task a short plan to the goal can be computed using only \ed{the fastest model, the model evaluation rate for our method is still high. The one-queue planner that uses MDEs is slower because the model precondition for the simulator model for \textit{LiftAndDrop} is usually satisfied, so some actions that are unnecessary to find a plan are simulated.}

The planning speeds for both methods using MDEs in \texttt{RodInDrawer} (Table~\ref{tab:timingOpenDrawer}) are faster than the simulator-only baseline because it can always use an analytical model for \textit{Pick} and \textit{OpenDrawer}, and sometimes use the analytical model for \textit{LiftAndDrop}. The improvement of using PE in addition to PS in this domain is smaller than for \texttt{RodInBox}, which we discuss further in the next paragraph. The random baseline is the slowest because it uses the largest number of simulator calls. 

Both MDEs and the planning method can affect the distribution of models chosen. For example, Figure~\ref{fig:modelsused} (left) shows planning with only PS causes the planner to compute successors using transitions in the model precondition of more expensive models at the same priority of sets in the faster models, which causes some actions to be simulated that would not be if they were using our planner using PS and PE, causing the longer planning times for \texttt{RodInBox}. For \texttt{RodInDrawer}, the planner using only PS selects very similar models to the planner using PS and PE, so using PE does not significantly affect planning speed. The reason is that the simulator model is more inaccurate in \texttt{RodInDrawer} than in \texttt{RodInBox}. The MDE can detect the task indirectly since the parameter distributions for the two tasks are different. As a result, the PS-only planner rarely uses the simulator, which can also be seen in Figure~\ref{fig:modelsused} (right). 
\newcommand{\modelselectedheight}{2.05cm}
\begin{figure}
    \centering
    \ed{
    \includegraphics[height=\modelselectedheight]{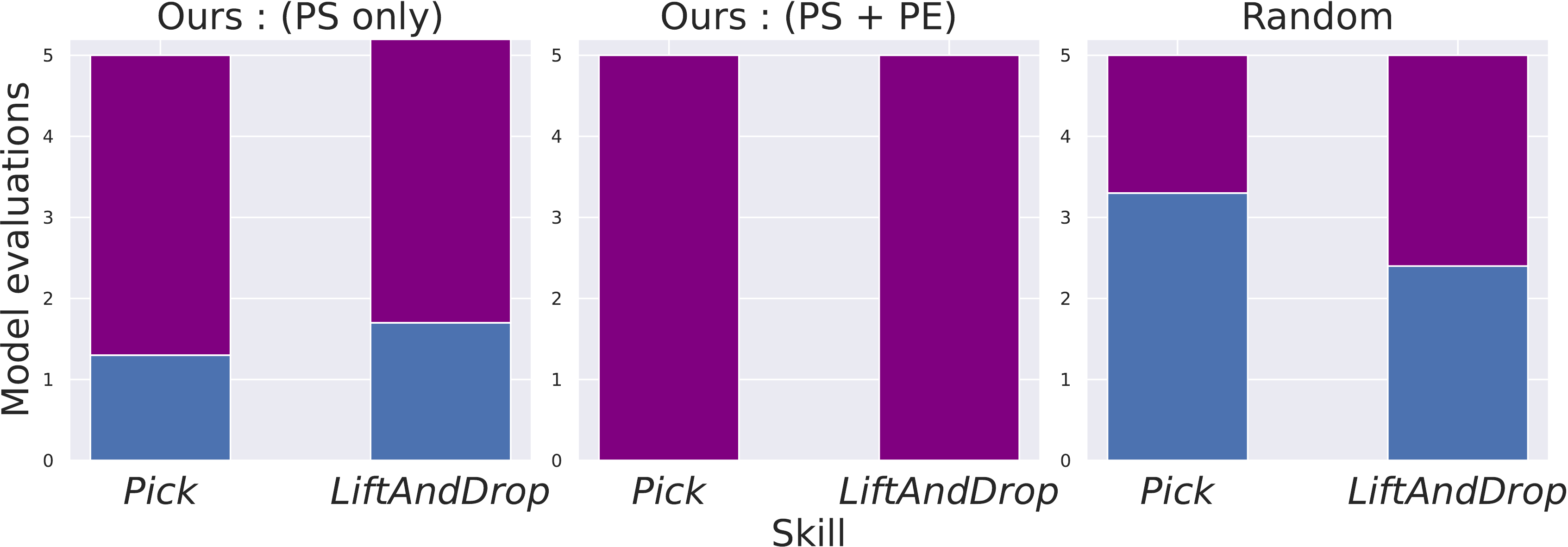}%
    \includegraphics[height=\modelselectedheight]{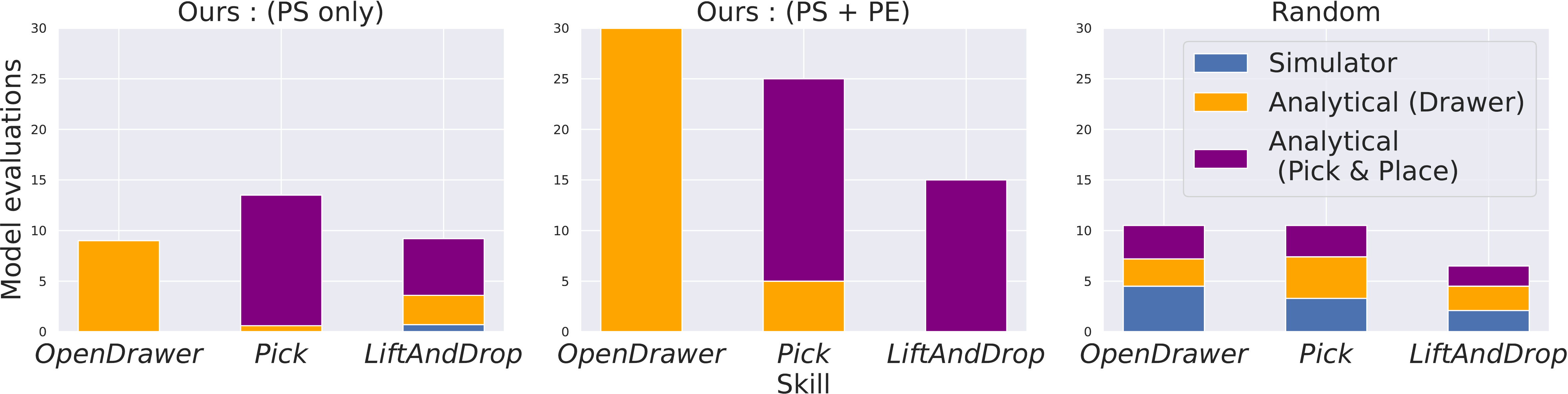}
    \caption{\footnotesize Average number of models evaluation of each type for all three skills for \texttt{RodInBox} (left) and \texttt{RodInDrawer}(right) across methods that use multiple models.}
    \label{fig:modelsused}
    }
    \vspace{-0.0cm}
\end{figure}

\reducesubtitlespace
\subsection{Reliability of Computed Plans}
\begin{figure}
    \centering
    \includegraphics[width=0.42\textwidth]{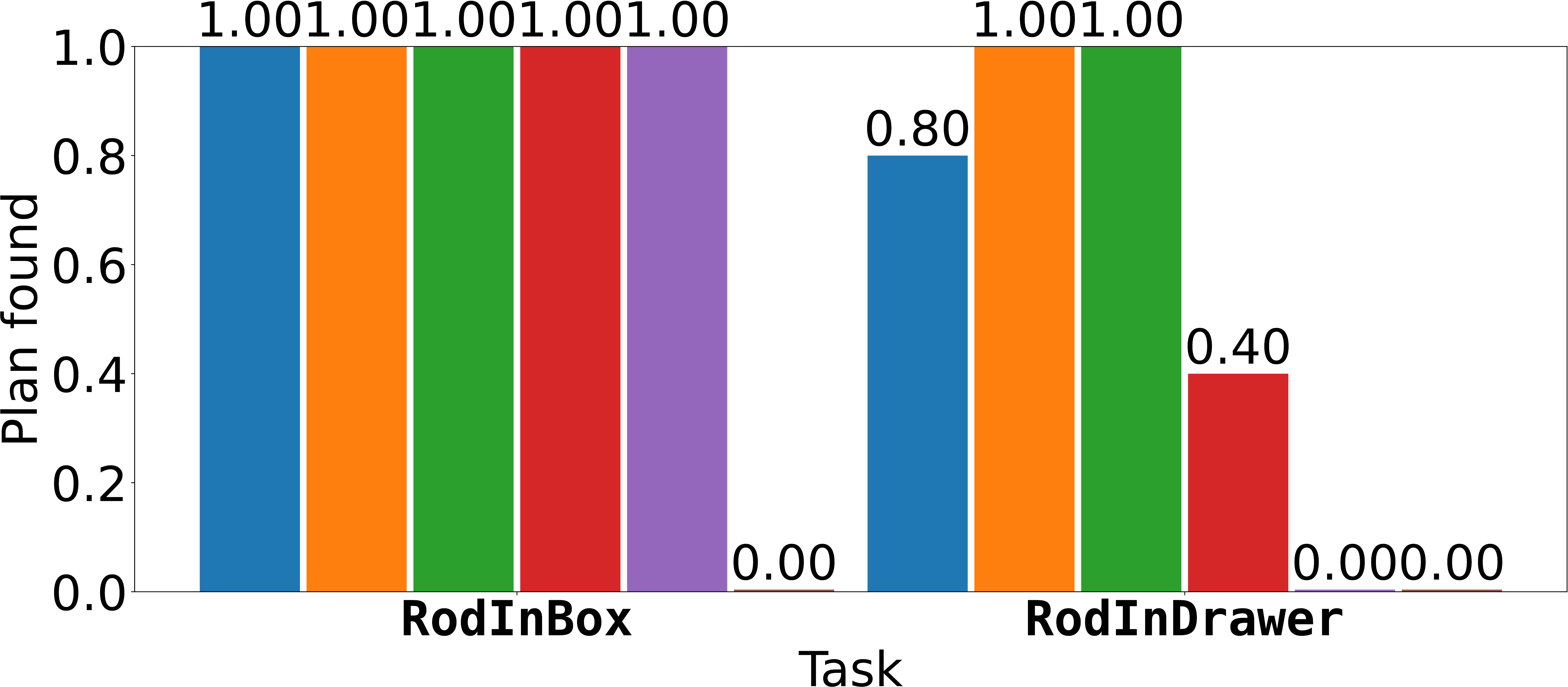}  
    \includegraphics[width=0.15\textwidth]{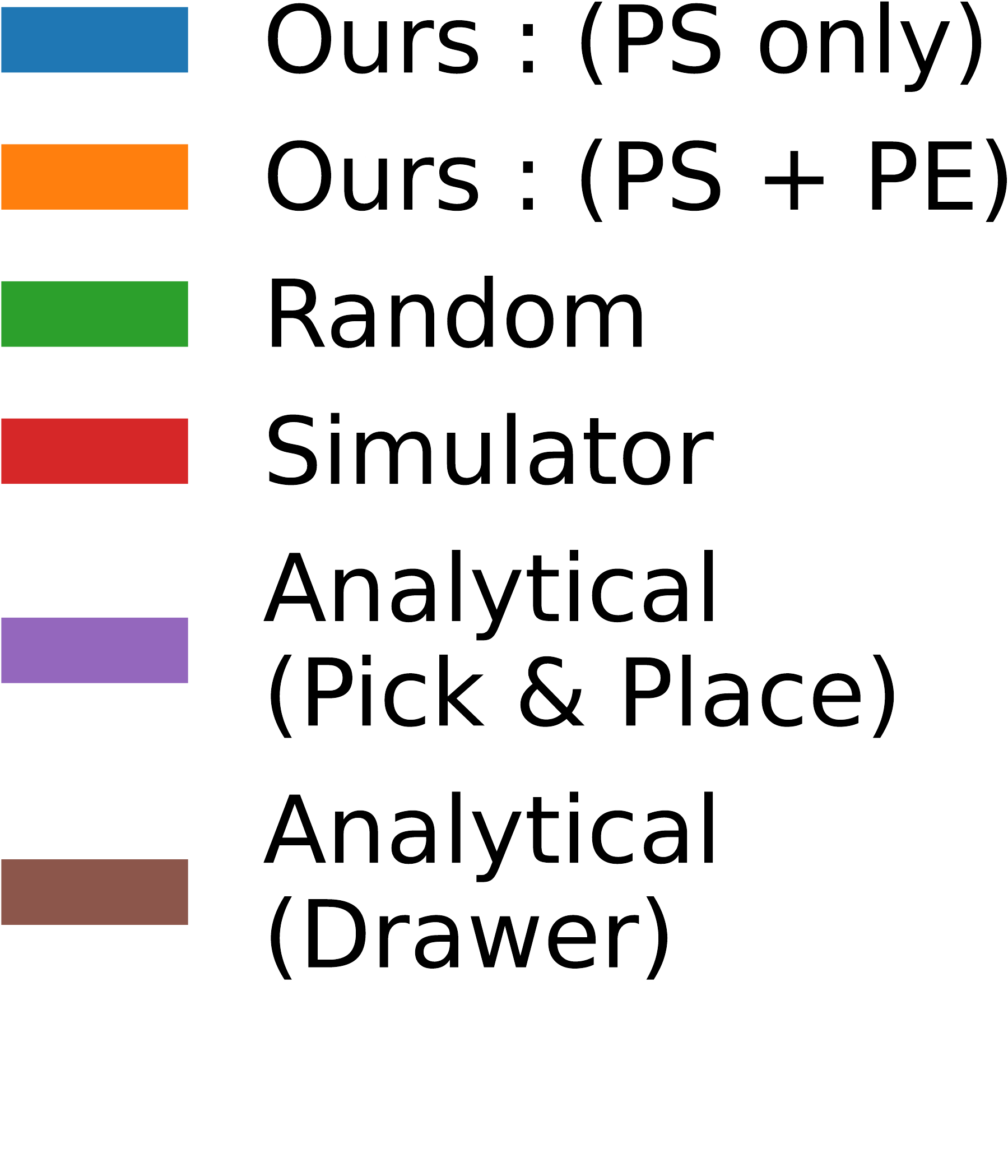}%
    \includegraphics[width=0.42\textwidth]{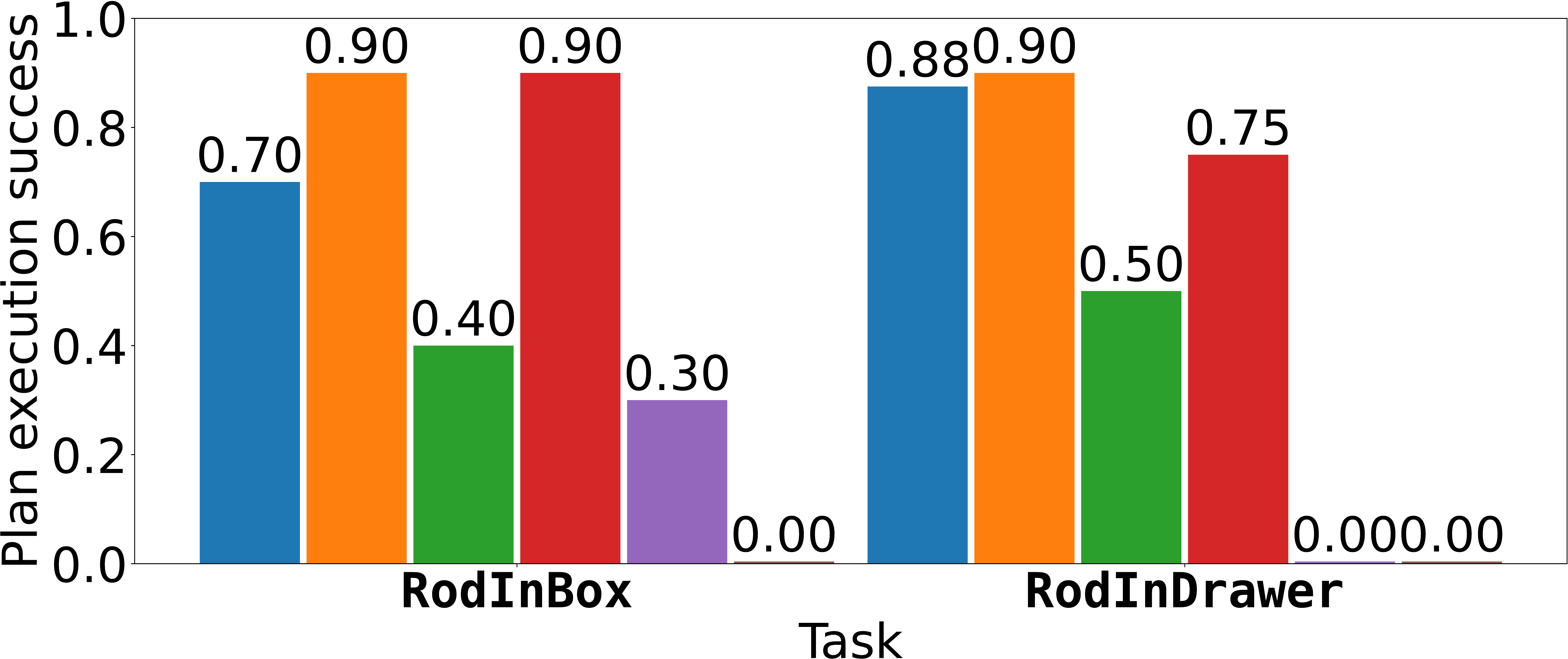}%
    \caption{\footnotesize Success rate in finding plans (left) and \ed{if found,} executing them until the goal (right).}
    \label{fig:plansuccess}
\end{figure}

\reducesubtitlespace
The experiments shown in Figure~\ref{fig:plansuccess}, evaluate the ability of our method to improve plan reliability without significantly increasing planning time by balancing model accuracy and computational cost.


 \begin{figure}[H]
     \centering
    \subfigandcaption{
    No pivot (Analytical)}
    {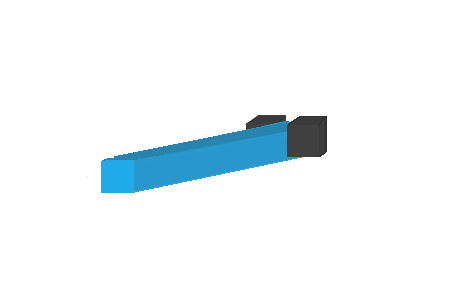}{0.245} \medskip %
    \subfigandcaption{
    Drop during transport (Real)}
    {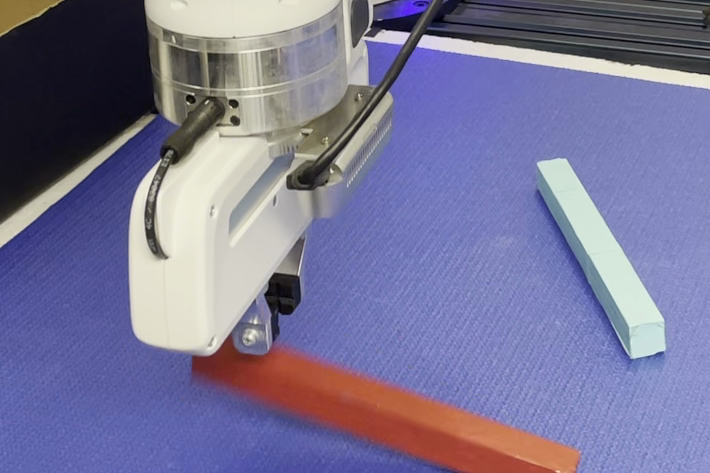}{0.245} \medskip %
    \subfigandcaption{Hitting target container (Real)}
    {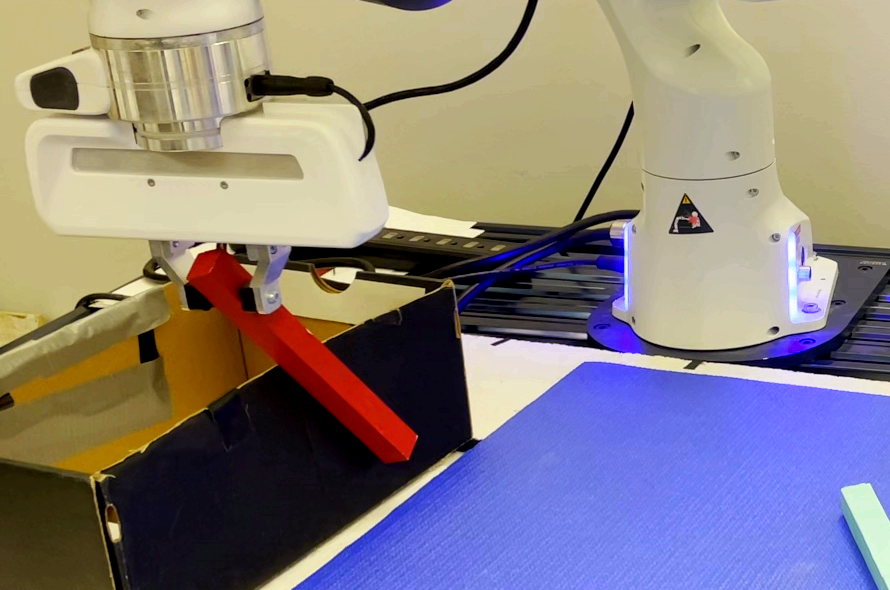}
    {0.245}\medskip\hspace{1pt}%
    \subfigandcaption{Unrealistic physics (Sim)}
    {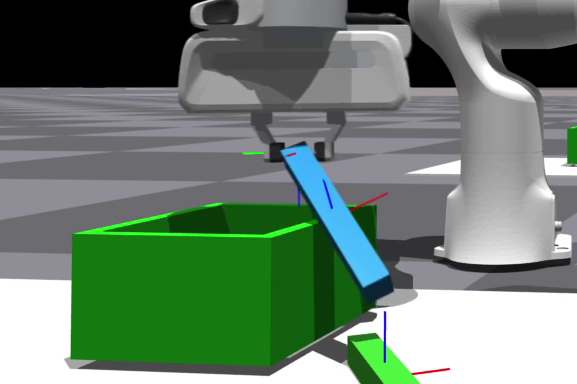}
    {0.245}%
     \caption{\footnotesize Situations where planning models are inaccurate. (a) \robotandrods assumes rigid attachment, although this (b) causes the rod to either drop during transport or (c) rotate and hit the target container. (d) The simulator sometimes exhibits unrealistic physics such as the rod sliding over the box walls.
     }
     \label{fig:failuremodes}
 \end{figure}

 \ed{All methods that use a model that reasons about interactions between the rod and robot are able to compute a plan in the allotted time for \texttt{RodInBox}. In the more challenging \texttt{RodInDrawer} task, the simulator-only baseline and one-queue version of our method sometimes fail to find a plan within the time limit.} 
 Plan success is highest for both methods using model preconditions and the simulator-only baseline. MDEs cause the planner to only use transitions where the model is predicted to be accurate, such as using \textit{LiftAndDrop} when the rod is grasped in the center. The simulator is more accurate than the analytical models so plans using it often succeed.
 In contrast, plans found using the analytical-model-only and random-model baselines often fail because the lowest cost paths often include grasping the end of a rod, which is predicted to be rigidly attached by the analytical model (Figure~\ref{fig:failuremodes}a).
 In the real world, grasping the end of the rod causes it to pivot, which usually causes the rod to drop during transport~(Figure~\ref{fig:failuremodes}b) or after hitting the target container (Figure~\ref{fig:failuremodes}c).
 Although the simulator can model pivoting in the gripper, unrealistic dynamics sometimes occur (Figure~\ref{fig:failuremodes}d).


\vspace{-0.3cm}
\section{Conclusion}
\vspace{-0.3cm}
\label{sec:conclusion}
We present a method for defining model preconditions using MDEs that predict model deviation given a state and a parameterized skill, \ed{which can be used to inform tradeoffs between models when each has complementary advantages and disadvantages in evaluation time and accuracy. Although MDEs can be used as constraints, we show a planner that uses multiple queues corresponding to different models that prioritizes planning using the faster models. Experimental evaluations show a speed up in planning by choosing between multiple models while keeping high plan reliability during execution. In both evaluation tasks, our results show that a robot can learn and reason about which transitions need to use more expensive models, which transitions none of the models can evaluate accurately, and how to inform planning with that information. For placing a rod in a drawer, we show that the robot can combine two low-fidelity models intended for representing different interactions, while avoiding using an expensive simulator that models all interactions. For future work, we will modify our framework to share more data across skills by using a shared action representation, which will also allow changing a low-level model multiple times during a skill execution. Furthermore, we will evaluate our method on tasks where model selection is more difficult, such as liquid, paper, and other deformable object manipulation. }



\clearpage


\section*{Acknowledgements}
This work was supported by the Office of Naval Research Grant No. N00014-18-1-2775, Army Research Laboratory grant W911NF-18-2-0218 as part of the A2I2 program, and National Science Foundation Grant No. CMMI-1925130 and IIS-1956163. Any opinions, findings,
and conclusions or recommendations expressed in this material are those of
the authors and do not necessarily reflect the views of our sponsors. The authors also thank Kevin Zhang for assistance with robot experiments, as well as Shivam Vats, Muhammad Suhail Saleem, Jacky Liang, Mohit Sharma, and Saumya Saxena for their support and discussions.
\bibliography{modelPreconditions}  

\begin{thebibliography}{30}
\providecommand{\natexlab}[1]{#1}
\providecommand{\url}[1]{\texttt{#1}}
\expandafter\ifx\csname urlstyle\endcsname\relax
  \providecommand{\doi}[1]{doi: #1}\else
  \providecommand{\doi}{doi: \begingroup \urlstyle{rm}\Url}\fi

\bibitem[Goyal et~al.(1991)Goyal, Ruina, and Papadopoulos]{goyal1991planar}
S.~Goyal, A.~Ruina, and J.~Papadopoulos.
\newblock Planar sliding with dry friction part 1. limit surface and moment
  function.
\newblock \emph{Wear (Amsterdam, Netherlands)}, 143\penalty0 (2):\penalty0
  307--330, 1991.

\bibitem[Pan and Hauser(2020)]{pan2020decision}
Z.~Pan and K.~Hauser.
\newblock Decision making in joint push-grasp action space for large-scale
  object sorting.
\newblock \emph{arXiv preprint arXiv:2010.10064}, 2020.

\bibitem[Liang et~al.(2018)Liang, Makoviychuk, Handa, Chentanez, Macklin, and
  Fox]{liang2018gpu}
J.~Liang, V.~Makoviychuk, A.~Handa, N.~Chentanez, M.~Macklin, and D.~Fox.
\newblock {GPU}-accelerated robotic simulation for distributed reinforcement
  learning.
\newblock In \emph{Conference on Robot Learning}, pages 270--282. PMLR, 2018.

\bibitem[Coumans and Bai(2016--2021)]{coumans2021}
E.~Coumans and Y.~Bai.
\newblock {PyBullet}, a python module for physics simulation for games,
  robotics and machine learning.
\newblock \url{http://pybullet.org}, 2016--2021.

\bibitem[Hafner et~al.(2019)Hafner, Lillicrap, Fischer, Villegas, Ha, Lee, and
  Davidson]{hafner2019learning}
D.~Hafner, T.~Lillicrap, I.~Fischer, R.~Villegas, D.~Ha, H.~Lee, and
  J.~Davidson.
\newblock Learning latent dynamics for planning from pixels.
\newblock In \emph{International Conference on Machine Learning}, pages
  2555--2565. PMLR, 2019.

\bibitem[Sundaralingam and Hermans(2021)]{sundaralingam2021hand}
B.~Sundaralingam and T.~Hermans.
\newblock In-hand object-dynamics inference using tactile fingertips.
\newblock \emph{IEEE Transactions on Robotics}, 2021.

\bibitem[Nagabandi et~al.(2020)Nagabandi, Konolige, Levine, and
  Kumar]{nagabandi2020deep}
A.~Nagabandi, K.~Konolige, S.~Levine, and V.~Kumar.
\newblock Deep dynamics models for learning dexterous manipulation.
\newblock In \emph{Conference on Robot Learning}, pages 1101--1112. PMLR, 2020.

\bibitem[Macklin et~al.(2014)Macklin, M{\"u}ller, Chentanez, and
  Kim]{macklin2014unified}
M.~Macklin, M.~M{\"u}ller, N.~Chentanez, and T.-Y. Kim.
\newblock Unified particle physics for real-time applications.
\newblock \emph{ACM Transactions on Graphics (TOG)}, 33\penalty0 (4):\penalty0
  1--12, 2014.

\bibitem[Kim et~al.(2019)Kim, Wang, Kaelbling, and
  Lozano-P{\'e}rez]{kim2019learning}
B.~Kim, Z.~Wang, L.~P. Kaelbling, and T.~Lozano-P{\'e}rez.
\newblock Learning to guide task and motion planning using score-space
  representation.
\newblock \emph{The International Journal of Robotics Research}, 38\penalty0
  (7):\penalty0 793--812, 2019.

\bibitem[Phillips et~al.(2012)Phillips, Cohen, Chitta, and
  Likhachev]{phillips2012graphs}
M.~Phillips, B.~J. Cohen, S.~Chitta, and M.~Likhachev.
\newblock E-graphs: Bootstrapping planning with experience graphs.
\newblock In \emph{Robotics: Science and Systems}, volume~5, page 110, 2012.

\bibitem[Konidaris et~al.(2018)Konidaris, Kaelbling, and
  Lozano-Perez]{konidaris2018from}
G.~Konidaris, L.~P. Kaelbling, and T.~Lozano-Perez.
\newblock From skills to symbols: Learning symbolic representations for
  abstract high-level planning.
\newblock \emph{Journal of Artificial Intelligence Research}, 61:\penalty0
  215--289, 2018.

\bibitem[Sharma and Kroemer(2020)]{Sharma-2020-126757}
M.~Sharma and O.~Kroemer.
\newblock Relational learning for skill preconditions.
\newblock In \emph{Proceedings of (CoRL) Conference on Robot Learning},
  November 2020.

\bibitem[Wang et~al.(2021)Wang, Garrett, Kaelbling, and
  Lozano-P{\'e}rez]{wang2021learning}
Z.~Wang, C.~R. Garrett, L.~P. Kaelbling, and T.~Lozano-P{\'e}rez.
\newblock Learning compositional models of robot skills for task and motion
  planning.
\newblock \emph{The International Journal of Robotics Research}, 40\penalty0
  (6-7):\penalty0 866--894, 2021.

\bibitem[Chou et~al.(2021)Chou, Ozay, and Berenson]{Chou2021ModelEP}
G.~Chou, N.~Ozay, and D.~Berenson.
\newblock Model error propagation via learned contraction metrics for safe
  feedback motion planning of unknown systems.
\newblock \emph{ArXiv}, abs/2104.08695, 2021.

\bibitem[Knuth et~al.(2021)Knuth, Chou, Ozay, and Berenson]{knuth2021planning}
C.~Knuth, G.~Chou, N.~Ozay, and D.~Berenson.
\newblock Planning with learned dynamics: Probabilistic guarantees on safety
  and reachability via {Lipschitz} constants.
\newblock \emph{IEEE Robotics and Automation Letters}, 6\penalty0 (3):\penalty0
  5129--5136, 2021.

\bibitem[Vemula et~al.(2020{\natexlab{a}})Vemula, Oza, Bagnell, and
  Likhachev]{Vemula-2020-122905}
A.~Vemula, Y.~Oza, J.~A. Bagnell, and M.~Likhachev.
\newblock Planning and execution using inaccurate models with provable
  guarantees.
\newblock In \emph{Proceedings of Robotics: Science and Systems (RSS '20)},
  July 2020{\natexlab{a}}.

\bibitem[Vemula et~al.(2020{\natexlab{b}})Vemula, Bagnell, and
  Likhachev]{vemula2020cmax++}
A.~Vemula, J.~A. Bagnell, and M.~Likhachev.
\newblock {CMAX}++: Leveraging experience for planning and execution using
  inaccurate models.
\newblock \emph{arXiv preprint arXiv:2009.09942}, 2020{\natexlab{b}}.

\bibitem[McConachie et~al.(2020)McConachie, Power, Mitrano, and
  Berenson]{mcconachie2020learning}
D.~McConachie, T.~Power, P.~Mitrano, and D.~Berenson.
\newblock Learning when to trust a dynamics model for planning in reduced state
  spaces.
\newblock \emph{IEEE Robotics and Automation Letters}, 5\penalty0 (2):\penalty0
  3540--3547, 2020.

\bibitem[Power and Berenson(2021)]{power2021keep}
T.~Power and D.~Berenson.
\newblock Keep it simple: Data-efficient learning for controlling complex
  systems with simple models.
\newblock \emph{IEEE Robotics and Automation Letters}, 6\penalty0 (2):\penalty0
  1184--1191, 2021.

\bibitem[Mitrano et~al.(2021)Mitrano, McConachie, and
  Berenson]{Mitranoeabd8170}
P.~Mitrano, D.~McConachie, and D.~Berenson.
\newblock Learning where to trust unreliable models in an unstructured world
  for deformable object manipulation.
\newblock \emph{Science Robotics}, 6\penalty0 (54), 2021.
\newblock \doi{10.1126/scirobotics.abd8170}.
\newblock URL \url{https://robotics.sciencemag.org/content/6/54/eabd8170}.

\bibitem[McConachie and Berenson(2020)]{McConachie2020bandit}
D.~McConachie and D.~Berenson.
\newblock \emph{Bandit-Based Model Selection for Deformable Object
  Manipulation}, pages 704--719.
\newblock Springer International Publishing, Cham, 2020.
\newblock ISBN 978-3-030-43089-4.
\newblock \doi{10.1007/978-3-030-43089-4_45}.
\newblock URL \url{https://doi.org/10.1007/978-3-030-43089-4_45}.

\bibitem[Saleem and Likhachev(2020)]{saleem2020planning}
M.~S. Saleem and M.~Likhachev.
\newblock Planning with selective physics-based simulation for manipulation
  among movable objects.
\newblock In \emph{2020 IEEE International Conference on Robotics and
  Automation (ICRA)}, pages 6752--6758. IEEE, 2020.

\bibitem[Youakim et~al.(2018)Youakim, Dornbush, Likhachev, and
  Ridao]{youakim2018motion}
D.~Youakim, A.~Dornbush, M.~Likhachev, and P.~Ridao.
\newblock Motion planning for an underwater mobile manipulator by exploiting
  loose coupling.
\newblock In \emph{2018 IEEE/RSJ International Conference on Intelligent Robots
  and Systems (IROS)}, pages 7164--7171. IEEE, 2018.

\bibitem[Du et~al.(2020)Du, Islam, and Likhachev]{du2020multi}
W.~Du, F.~Islam, and M.~Likhachev.
\newblock Multi-resolution a.
\newblock In \emph{Thirteenth Annual Symposium on Combinatorial Search}, 2020.

\bibitem[Sutton et~al.(1999)Sutton, Precup, and Singh]{sutton1999between}
R.~S. Sutton, D.~Precup, and S.~Singh.
\newblock Between {MDPs} and semi-{MDP}s: A framework for temporal abstraction
  in reinforcement learning.
\newblock \emph{Artificial intelligence}, 112\penalty0 (1-2):\penalty0
  181--211, 1999.

\bibitem[Aine et~al.(2016)Aine, Swaminathan, Narayanan, Hwang, and
  Likhachev]{aine2016multi}
S.~Aine, S.~Swaminathan, V.~Narayanan, V.~Hwang, and M.~Likhachev.
\newblock Multi-heuristic {A*}.
\newblock \emph{The International Journal of Robotics Research}, 35\penalty0
  (1-3):\penalty0 224--243, 2016.

\bibitem[Zhang et~al.(2020)Zhang, Sharma, Liang, and Kroemer]{zhang2020modular}
K.~Zhang, M.~Sharma, J.~Liang, and O.~Kroemer.
\newblock A modular robotic arm control stack for research: Franka-interface
  and frankapy.
\newblock \emph{arXiv preprint arXiv:2011.02398}, 2020.

\bibitem[Garrett(2020)]{pybulletplanning}
C.~R. Garrett.
\newblock Pybullet planning, 2020.
\newblock URL \url{https://pypi.org/project/pybullet-planning}.

\bibitem[Kingma and Ba(2015)]{kingma2014adam}
D.~P. Kingma and J.~Ba.
\newblock Adam: {A} method for stochastic optimization.
\newblock In Y.~Bengio and Y.~LeCun, editors, \emph{3rd International
  Conference on Learning Representations, {ICLR} 2015, San Diego, CA, USA, May
  7-9, 2015, Conference Track Proceedings}, 2015.
\newblock URL \url{http://arxiv.org/abs/1412.6980}.

\bibitem[Likhachev et~al.(2003)Likhachev, Gordon, and Thrun]{likhachev2003ara}
M.~Likhachev, G.~J. Gordon, and S.~Thrun.
\newblock Ara: formal analysis.
\newblock 2003.

\end{thebibliography}
\clearpage
\section{Appendix}
\subsection{Skills}
In this section, we describe the skills used, their preconditions, parameter generation methods, and  termination conditions. The rod width is $w$, length is $l$, and height is $h$.

A common precondition for all skills is that any desired pose is within reachable ranges of the robot of the robots workspace where control works well, which are specified as a $0.5 m \times 0.6$ rectangle in $x$ and $y$.

Low-level motion plans for \textit{Pick} and \textit{LiftAndDrop} move in 2D in an obstacle free $z_{\text{free}}$ region to get to a target pose. \textit{OpenDrawer} also moves in $z_{\text{free}}$ until it needs to open the drawer.

Some common hyperparameters:
\begin{enumerate}
    \item $\epsilon_{gripper}$: An amount beyond $w$ to close the gripper, which should be very small or the skill will not terminate. 
    \item $\epsilon_p$: Maximum position error for $[x,y,z]$
    \item $\epsilon_{\psi}$:  Maximum yaw error. 
    \item $w_{\text{open}}$ : The width at which the gripper is considered open, which is wider than $w$.
\end{enumerate}
\label{sec:skill_descs}
\subsubsection{Pick}
This skill moves the robot to a grasp around an object. The parameters are $[x_d,y_d,z_d,\psi_d]$ which represent the desired final pose of the end-effector. We assume rods are grasped at their narrowest axis so the gripper width is $w - \epsilon_g$

The precondition is that no rod is within $l/2$ of the robot gripper, and the open gripper position will not cause collisions with other rods. The termination condition is that the end-effector pose is close to the desired pose: $||[x_d,y_d,z_d] - [x,y,z]|| < \epsilon_p$, $|\psi_d - \psi| < \epsilon_{\psi}$ and the gripper width is less than some $w-\epsilon_g$

Pick is relatively easy to model with a kinematic model, though failures can occur from collision with the drawer not modeled in the analytical model in \texttt{RodInDrawer}, reaching joint limits, or perception error. In our setup with our parameter generators, these errors were rare, but would not be ruled out by the skill precondition. 

Grasps around the center of the rod, and a fixed distance from the ends are used to compute $[x_d,y_d,z_d, \psi_d]$. 

\subsubsection{\textit{LiftAndDrop}}
\textit{LiftAndDrop} lifts the rod from the ground, goes to $[x_d,y_d,z_d,\psi_d]$, and then opens the gripper, allowing the rod to drop.

The precondition is that a rod is within $l/2$ of the gripper. The termination condition, like \textit{Pick}, requires $||[x_d,y_d,z_d] - [x,y,z]|| < \epsilon_p$, $|\psi_d - \psi| < \epsilon_{\psi}$ and that the gripper width is greater than $w_{\text{open}}$.

Task-specific parameters $[x_d, y_d, z_d]$ are sampled within a rectangle defined by the goal (box or drawer) at a fixed height above the object. The rectangle is some fraction of the goal region, which was conservatively set at 2 cm.

\textit{LiftAndDrop} is a placement task, which is much harder to predict, especially at a height because the rod interacts with the ground, goal box, and goal drawer, potentially bouncing unpredictably. Some analytical model failures result from hitting the walls of the box while grasping a rod from the side, and dropping the rod on the top of the chest holding the drawer. The simulator model is fairly accurate though it does not perfectly model all interactions with the drawer.

\subsubsection{\textit{OpenDrawer}} 
For this skill, a robot opens a drawer by placing the gripper parallel to the front edge of the drawer and then moves the gripper to a second goal pose along the drawer axis. This skill assumes the drawer is opened along the negative $y$ axis in world frame. The parameters are $[x_d, y_d, z_d,\psi_d, y_{\text{open}}]$ The first goal pose is $[x_d,y_{d,i}, z_d, \psi_d]$. The robot pushes the drawer by moving along the negative $y$ axis to a final $y_{d,f}$ The precondition is that there is no rod already in the hand, and that the drawer is not already open. The parameters are $[x_d,y_{d,i},z_d,\psi_d,y_{d,f}]$. Parameter generation is object-centric, sampled based on the drawer pose to a random opened amount.

The precondition for \textit{OpenDrawer} is that the gripper does not have any objects within $l/2$ of its starting pose.

The termination condition is reached once the robot has reached the final pose at $[x_d, y_d-y_{\text{open}}, z_d, \psi_d]$. 

Parameters are sampled by selecting $[x_d, y_d, z_d]$ so the robot is a fixed distance (2 cm) behind the front of the drawer in y, 1 cm offset from the center in $x$, and 4 cm above the drawer bottom in $z$.  and $\psi_d$ is selected to the gripper is parallel to the long side of the front of the drawer. $y_{\text{open}}$ was randomly sampled from a small range: [14,17] cm. 

\subsection{Skill-Level Models}
\subsubsection{Simulator Models}
\label{sec:simparams}
We use a simulator to model complex interactions with rods, target objects, and the robot. As a result, the simulator should be accurate in at least some contexts to be useful. It is non-trivial to tune the parameters to get realistic behavior with small objects that rotate in-hand and behave realistically with thin objects such as drawer walls or box walls, so simulator parameter tuning is important. The parameters used for each IsaacGym simulator that do not match default parameters are shown in Table~\ref{tab:simparams}. We also show the joint properties and gains used for the Franka below the simulator properties. Lastly we show simulator environment properties. The friction coefficient between the rods and other objects in the scene including the gripper is sampled from a small range: $[0.50, 0.51]$. The torsional friction coefficient is $1 \times 10^{-13}$. The mass is sampled from the range $[50, 51]$ g. and the dimensions are the same as they are in the real world: 2.3 x 18.5 x 2.3 cm. The drawer is modeled as a prismatic joint with approximated dynamics and limits from visual inspection.

Because the state in our experiments does not include joint angles, the robot state for the simulator model is underconstrained. To solve this problem, we use inverse-kinematics, optimizing for an upright elbow using the tools provided in~\cite{pybulletplanning}.

\begin{table}[H]
    \centering
    \begin{tabular}{l|l}
       Property & Value \\ 
       \hline
       engine & physx \\
       dt & $2.5 \times 10^{-3}$ for \texttt{RodInBox} and $5 \times 10^{-3}$ for \texttt{RodInDrawer} \\
       num\_position\_iterations  &  8  \\
       rest\_offset  & 0 \\
       contact\_offset & $1 \times 10^{-3}$ \\
       friction\_offset\_threshold & $1 \times 10^{-3}$ \\
       friction\_correlation\_distance & $5 \times 10^{-4}$ \\
        \hdashline
        stiffness & 1000\\
        damping & 500 \\
        \hdashline
        joint stiffness & [0, 0, 0, 0, 0, 0, 0, $1 \times 10^{3}$, $1 \times 10^{3}$] \\
        joint damping &[0, 0, 0, 0, 0, 0, 0, 200, 200]  \\
        effort & [0, 0, 0, 0, 0, 0, 0, 10, 10]
    \end{tabular}
    \vspace{0.3cm}
    \caption{Simulation parameters for environment including physics simulator properties, attractor properties, and joint properties.}
    \label{tab:simparams}
\end{table}

\subsubsection{Analytical Models}
\label{sec:analyticalmodels}
When there is no object in the gripper, \robotandrods predicts that the robot goes to the commanded goal pose described by the action parameters, $\tf{ee}{goal}$, and all other objects do not move. 
If there is an object close to the gripper, the transform from robot gripper to object, $\tf{rod}{ee}$, is computed from the initial state: $\tf{rod}{ee} = \tf{ee}{world}^{-1} \cdot \tf{rod}{world}$, using the closest rod to the gripper as the target rod.
The end effector is assumed to move to the desired pose then drop the rod to the ground so the new pose of the rod is computed using the initial $\tf{rod}{ee}$ by applying the transform to the end-effector at the goal pose. Let $h_{\mathrm{rod}}$ be the height of the rod, and $h_{\mathrm{drawer}}$ the height of the drawer at the rod's final pose. 
The final $z$ value after the rod is dropped is $\frac{h_{\mathrm{rod}}}{2}$ if on the floor, and $h_{\mathrm{drawer}} + \frac{h_{\mathrm{rod}}}{2}$ if the rod is dropped over the drawer. 

\robotanddrawer uses the geometry of both the drawer and the grippers to compute whether at least one gripper is in contact with the drawer edge. If they are in contact, then the drawer moves the amount specified in the skill parameters. Otherwise the drawer does not move. This model does not account for the joint limits of the drawer. 

\subsection{MDE Training}
\label{mdetrainingdetails}
In this section, we describe how we train the MDEs. As mentioned in Section~\ref{sec:learningmodelpreconds}, training data is collected by computing plans using all models, and then executing them. We collected 26 trajectories solving \texttt{RodInBox} and 17 trajectories solving \texttt{RodInDrawer}. 15\% of samples were held out for testing. Since real-robot data is expensive to collect, data augmentation is critical to good MDE performance. In addition to doubling the dataset by swapping the two rod locations, we also added random noise from a normal distribution with standard deviation 1 cm for state features and 3 cm for parameter features. Of the augmented dataset, 5\% were removed from the training set to use for validation which informed early stopping to prevent overfitting. 
The MLP hyperparameters are shown in Table~\ref{tab:mlpparams}. The optimizer we use is the Adam optimizer \citep{kingma2014adam}.

For all models,  $c_1=3$ (underestimate loss coefficient) and $c_2 = 1$ (overestimate loss coefficient).

\begin{table}[H]
    \centering
    \begin{tabular}{l|l}
        Hyperparameter & Value  \\
        \hline
         Number layers & 2 \\
         Activation & ReLU \\
         Learning rate & $5 \times 10^{-3}$ \\
         Hidden units per layer & 64 \\
         Adam L2 weight decay & $5 \times 10^{-3}$ 
    \end{tabular}
    \caption{MLP hyperparameters}
    \label{tab:mlpparams}
\end{table}

\subsection{Planning Algorithm}
\label{sec:planningalg}
Here, we describe more details for the planner, including hyperparameters, heuristics, guarantees and pseudocode. 

The planner hyperparameters for \texttt{RodInBox} are $w=[10,1]$, $\epsilon=5$ (for WA*). For \texttt{RodInDrawer} the planner hyperparameters are $w=[10,1.1,1]$, $\epsilon=10$ 

The heuristic we use is task dependent. For \texttt{RodInBox}, the heuristic is the distance between the target rod and the center of the box. For \texttt{RodInDrawer}, the heuristic is the distance between the target rod and center of the box, plus the distance between the current drawer location and the minimum required to open it to place a rod in. 
\subsubsection{Algorithm pseudocode}
\newcommand{\sbest}{s_{\mathrm{best}}}
\begin{algorithm}[H]
    \caption{Planning using multiple queues given: \\
    start state $s_0$, \\
    goal set $\mathcal{S}^g$, \\
    skills that generate $a$ given $s$, \\
    list of models from slowest to fastest $[M_0, M_1, \ldots, M_K]$, and \\
    model cost weights $[w_0, w_1 \ldots w_K]$}
\ed{
    \begin{algorithmic}
        \Procedure{Search}{$s_0, \mathcal{S}^g, \mathbb{A}, [M_0, M_1, \ldots, M_K], \mathbf{w}$}
        \State Initialize start node with $s_0$ and $\aincomplete(s_0) \gets \emptyset$
        \State insert start node into open queues with $f \gets g+wh(s_0)$
        \While{any open queue is nonempty}
            \State $i \gets \underset{i<=K}{\argminC} \,  w_i\open_i.MinKey()$
            \State $ \sbest \gets Pop(\open_i.MinKey()) $
            \If{$\aincomplete(\sbest)$ is nonempty}
                \State $A \gets \aincomplete(\sbest)$
            \Else
                \State Generate $A$ using parameter generators from each skill which satisfy $\mathrm{pre}(a)$
            \EndIf
            \If{$\sbest \in \mathcal{S}^g$ }
                \State Return path from node 
            \EndIf
            \If{$i = 0$}
                \State $successors \gets \Call{FullExpansion}{\sbest}$
                \State Add $\sbest$ to all closed sets
            \Else
                \State $successors \gets \Call{PartialExpansion}{\sbest}$
                \State Add $\sbest$ to $\closed_i$
                 
            \EndIf
            \For {$s' \in successors$}
                \State $\aincomplete(s') \gets \emptyset$
                \For {$i \in \{0,K\} $} \Comment{Update all graphs}
                \If{$s' \notin \closed_i$ and $g(s') > g(s) + c(s, a, s') $}
                    \State $g(s') \gets g(s) + c(s, a, s')$
                    \State Add $s'$ to all open queues 
                \EndIf
                \EndFor
            \EndFor
        \EndWhile
        \EndProcedure
    \end{algorithmic}
    }
     \label{alg:planningAlg}
\end{algorithm}
\label{sec:pseudocode}

\begin{algorithm}[H]
    \caption{Full expansion of node}
    \label{alg:fullexpand}
    \ed{
    \begin{algorithmic}
    \Procedure{FullExpansion}{$s$}
       \State $successors \gets \emptyset$
       \For{$a \in A$}
          \If {$s \in \mathrm{pre}(a)$}
                \State Compute $s'$ using highest (fastest) $i$ such that $(s, a) \in \mathrm{pre}(M_i)$
                \State Add $s'$ to $successors$
          \EndIf
     \EndFor
     \State \Return $successors$
    \EndProcedure
    \end{algorithmic}}
\end{algorithm}

\begin{algorithm}[H]
    \caption{Partial expansion procedure for only computing successors in the precondition for $M_i$}
    \label{alg:partialexpand}
    \begin{algorithmic}
    \Procedure{PartialExpansion}{$s, i$}
     \State $successors \gets \emptyset$
       \For{$a \in A$}
            \If $(s,a) \in \mathrm{pre}(M_i)$ 
              \State Compute $s'$ using $M_i(s, a)$
              \State Add $s'$ to $successors$
             \Else
                \State Add $a$ to $\aincomplete(s)$
            \EndIf
      \EndFor
     \State return $successors$
    \EndProcedure

    \end{algorithmic}
\end{algorithm}
\subsubsection{Hyperparameter selection}
Next we show the $d_{\text{max}}$ parameters for each skill in Table~\ref{tab:dmax}. These currently are chosen by hand based on task accuracy needs. In future work, we plan to explore automating selecting $d_{\text{max}}$, as well as use a different $d$ to reflect the consequences of predicting an incorrect contact mode. Predicting the wrong contact mode can lead to very different dynamics despite Euclidean distances being very similar. 

\begin{table}[H]
    \centering
    \begin{tabular}{c|l| l}
        Task & Skill & $d_{\text{max}}$  \\
        \hline
        \multirow{2}*{\texttt{RodInBox}} & \textit{Pick} &  3 cm \\
         & \textit{LiftAndDrop} & 8 cm \\
         \hdashline
         \multirow{3}*{\texttt{RodInDrawer}}  & \textit{OpenDrawer} &  6 cm \\
          & \textit{Pick} & 3 cm \\
          & \textit{LiftAndDrop} & 5 cm\\
    \end{tabular}
    \caption{$d_{\text{max}}$ values used for each skill and task.}
    \label{tab:dmax}
\end{table}

\subsection{Suboptimality guarantees}
\label{sec:proof}
\newcommand{\wtwo}{\frac{w_0}{\minw}}
\newcommand{\minw}{\mathrm{min}(w)}
We prove that our algorithm provides suboptimality guarantees by showing it is a special case of Weighted A* when using PS and Multi-Resolution A* when using PE. We only consider edges executing $a$ from $s$ to reach $s'$ if it can be accurately modeled by at least one $M$. Otherwise, the edge does not exist. The guarantees are only on the plan; our method for learning $\mathrm{pre}(M)$ using MDEs does not have formal guarantees on the model precondition accuracy. 

\textbf{Assumptions: }
\begin{itemize}
    \item $M_i(s,a) = M_j(s,a) \forall (i,j) s.t. (s,a) \in \mathrm{pre}(M_i) \cap \mathrm{pre}(M_j)$, we know that all models will predict the same discrete state if correct 
    \item $\epsilon > 1$
    \item $w_i >= w_j$ if $i < j$, reflecting decreasing penalty for each model
    \item $h$ is admissible and consistent
    \item $c(s,s') \geq 0$
    \item If $(s,a)$ is not in $\mathrm{pre}(M)$ for any $M$ in the problem, any outgoing edge using $a$ from $s$ has infinite cost
\end{itemize}

\begin{theorem} If WA* with PS terminates, $g(s_g) \leq \epsilon g^*(s_g)$ \end{theorem}
\begin{proof}
Since the rule for expansion in PS is to select $\mathrm{max}_i s.t. s,a \in \mathrm{pre}(M_i)$, if $ s,a \in \bigcup \mathrm{pre}(M_i)$, then there will be at least model that can be used to compute $s'$. Otherwise, that edge does not exist. Because of Assumption 2, successors for any $s$ is the same whether computed using $M_i$ or $M_j$. Thus, if $\exists M s.t. (s,a) \in \mathrm{pre}(M)$, $s'$ will be computed. Since $c(s,a,s')$ is not dependent on $M$, the cost will be the same, so $g(s')$ is not dependent on the model. Therefore, using PS does not change the properties of WA*. 

Using the fact that WA* with PS is a special case of WA*, the cost to $s_g$ found using MRA* using PS is upper bounded by $\epsilon g^*(s_g)$. The bound follows from the proof in ~\citep{likhachev2003ara}. 
\end{proof}

\begin{theorem}{If WA* with PE terminates, $g(s_g) \leq \frac{w_0}{\minw}\epsilon g^*(s_g)$} \end{theorem}
\begin{proof}

Here, we show that $g(s_g) \leq \frac{w_1}{\minw} g^*(s_g)$ when using W* with PE by showing it is a special case of MRA*. Like MRA*, our algorithm uses an anchor search when $i=0$ and uses additional searches that use different action sets when $i>0$. 
There are no restrictions on the implementation of the procedure that selects which queue to use when $i \neq 1$ in MRA*. Thus, for $i > 0$, using Equation~\ref{eq:expansionrule} is a valid implementation of \textsc{ChooseQueue}.
MRA* also employs partial expansions to add successors that correspond with a given search. In our algorithm's partial expansions, the same mechanism is used but by only returning successors generated using model $M_i$. 
We complete the connection to MRA* by showing that $\wtwo$ is the variable used to prioritize inadmissible searches over the anchor search in our algorithm. 
The decision rule used for PE (Equation~\ref{eq:expansionrule}) can be written as the rule for whether to use the anchor queue or use an additional search in MRA* by rearranging terms from Equation~\ref{eq:expansionrule}: 
\begin{equation}
\wtwo \open_0 MinKey () \leq \open_i.MinKey() \forall i > 0
\label{eq:rewrittenpe}
\end{equation} 

Therefore, we can use the result that $g(s) \leq \alpha g^*(s) \forall s$ for some constant $\alpha$ to guarantee that $g(s_g) \leq \alpha g^*(s_g)$ where $\alpha = \wtwo \epsilon$ when using WA* with PE.

\end{proof}


\end{document}